\documentclass[journal]{IEEEtran}

\usepackage{graphicx}
\usepackage{amsmath}
\usepackage{amssymb}
\usepackage{booktabs}
\usepackage{array}
\usepackage{algorithm}
\usepackage{algpseudocode}
\usepackage{xcolor}
\usepackage{multirow}
\usepackage{siunitx}
\usepackage{placeins}
\usepackage[colorlinks=true, linkcolor=black, citecolor=black, urlcolor=blue]{hyperref}
\usepackage[scaled=0.92]{helvet}

\algnewcommand\algorithmicparfor{\textbf{for}}
\algnewcommand\algorithmicpardo{\textbf{do in parallel}}
\algdef{SE}[PARFOR]{ParFor}{EndParFor}[1]
  {\algorithmicparfor\ #1\ \algorithmicpardo}
  {\algorithmicend\ \algorithmicparfor}

\begin{document}
\bstctlcite{IEEEexample:BSTcontrol}

\title{
Multi-Agent Reinforcement Learning of Co-Adaptive Hip Exoskeleton Control (MARL-Exo) for Metabolic Energy Reduction
}

\author{Yifei~Yuan,
        Jakob~Wolf,
        Ghaith~Androwis,
        and~Xianlian~Zhou%
\thanks{Yifei Yuan, Jakob Wolf, and Xianlian Zhou are with the Department of Biomedical
Engineering, New Jersey Institute of Technology, Newark, NJ 07102, USA
(e-mail: yy72@njit.edu;jw768@njit.edu;alexzhou@njit.edu).}%
\thanks{Ghaith Androwis is with Kessler Foundation, West Orange, NJ 07052, USA
(e-mail: GAndrowis@kesslerfoundation.org).}%
\thanks{This work was partially supported by the National Institute on Disability, Independent Living, and Rehabilitation Research (NIDILRR) funded Rehabilitation Engineering Research Center Grant 90REGE0025-01-00 and NSF Award \#2524089. The human study was approved by the Institutional Review Board under Protocol No.\ 2305033091R003 and conducted in accordance with the Declaration of Helsinki.}%
\thanks{This work has been submitted to the IEEE for possible publication. Copyright may be transferred without notice, after which this version may no longer be accessible.}}

\markboth{}{}

\maketitle

\begin{abstract}\boldmath
\textit{Objective:} This study aims to determine whether a hip-exoskeleton policy trained in simulation, by co-adapting with a musculoskeletal human model under a metabolic-energy reward, reduces the metabolic cost of walking in new users without per-user tuning, and how mechanical-power shaping affects the learned assistance.
\textit{Methods:} We developed MARL-Exo, a multi-agent reinforcement learning framework in which a full-body human model (304 muscle actuators; randomized anatomy, gait, and speed) and a bilateral hip-exoskeleton policy are trained concurrently in two stages. The exoskeleton observes only delayed hip kinematics and its past commands; its reward includes muscle metabolic power from a physiologically based energetics model, alone (Metabolic) or with mechanical-power shaping (Metabolic\,+\,Power). The Metabolic policy was deployed and tested without retraining in eight healthy adults walking at 1.25\,m/s, using indirect calorimetry.
\textit{Results:} In simulation, the Metabolic policy reduced modeled muscle metabolic power by 13.8\% relative to no-device walking, versus 8.8\% for Metabolic\,+\,Power. On hardware, MARL-Exo reduced net metabolic rate by $18.3\pm6.6\%$ relative to Passive in all eight participants. Relative to No-Exo, the mean reduction was $7.4\pm10.0\%$; six participants more than offset the $13.1\%$ device-wearing penalty, walking up to $24.2\%$ below their No-Exo cost.
\textit{Conclusion:} A metabolically informed policy co-adapted in simulation transferred to users without per-user tuning and consistently reduced the metabolic cost of walking with the device, while explicit power shaping reduced the simulated benefit.
\textit{Significance:} Unlike per-user human-in-the-loop optimization, this approach yields a single population-level controller and allows assistance objectives to be screened in simulation before human testing.
\end{abstract}


\section{Introduction}
\IEEEPARstart{P}{owered} lower-limb exoskeletons are being developed both for gait rehabilitation and for augmenting healthy walking~\cite{rodriguez2021systematic,baud2021review}. Among them, hip exoskeletons are especially attractive because the hip musculature performs a large fraction of the positive work of level locomotion, so assistance there can offload muscles that do much of the work of walking. Delivering that benefit efficiently is challenging because the wearer is not a passive plant. Once a device applies torque at a joint, the mechanics of gait change, and the neuromuscular system responds by recruiting muscles differently and re-coordinating the limbs~\cite{poggensee2021adaptation}. Each side of the interaction thus changes the operating conditions of the other, and the controller is tuned against a wearer who is still adapting. The problem is compounded by the choice of objective: active devices can shape the assistance profile flexibly, but the benefit they deliver depends strongly on what the controller is optimized for, and on how well that objective reflects the wearer's response. Human-in-the-loop optimization tunes the objective directly on the wearer, but it demands long per-user experiments that must be repeated for every user and condition~\cite{zhang2017human,ding2018human}.

Learning the controller in musculoskeletal simulation offers a way around those experiments, since the wearer's response can be simulated rather than measured. A neural network control policy for a full lower limb exoskeleton coupled to a musculoskeletal model was first used to learn robust assistance for walking~\cite{luo2023robust}, and a similar idea was used for joint human--exoskeleton training with sim-to-real transfer that produced a control policy for a hip exoskeleton that lowered metabolic cost without subject-specific tuning~\cite{luo2024experiment}. The paradigm has since been extended to speed- and slope-aware assistance through muscle-synergy priors~\cite{park2026learning}, to reflex-based human models paired with end-to-end hip policies~\cite{barati2026end}, to human-aligned simulation of assistance parameters~\cite{leem2026exo}, and to asymmetric gait~\cite{yuan2026gait}, squatting~\cite{ratnakumar2026reinforcement}, sit-to-stand~\cite{ratnakumar2026predicting}, and running~\cite{simos2025reinforcement}. 

Two gaps remain in this literature. First, the mutual adaptation of human and device is mostly left implicit: when both policies are optimized at once, each shifts the data distribution seen by the other, and training can drift toward oscillatory torque, mistimed assistance, or divergence. Second, the effort terms used in training are typically proxies such as muscle activation or work. Luo \emph{et al.}~\cite{luo2024experiment} explicitly chose activation over a metabolic-rate term because of calibration and reward-smoothness concerns, and other controllers use kinematic tracking or activation-based effort~\cite{park2026learning,ratnakumar2026reinforcement}. What a wearer ultimately experiences is metabolic energy, which a musculoskeletal simulation can compute directly but which, to our knowledge, has not been used as a term of the exoskeleton policy reward. The closest examples are of two kinds. Exo-Plore rewards a simplified activation-weighted energy expression rather than a physiological muscle-energetics model~\cite{leem2026exo}. Barati \emph{et al.}~\cite{barati2026end} optimize their reflex-based human model with Umberger's energetics as the objective; however, this objective was not incorporated into the exoskeleton policy reward, which instead used an activation-based effort term.

Curriculum learning is a natural remedy for the first gap, since decomposing a hard learning problem into ordered stages improves stability~\cite{bengio2009curriculum} and physiological plausibility~\cite{chiappa2024acquiring,caggiano2022myosuite}. We previously proposed Staged Multi-Agent Training (SMAT), a four-stage curriculum trained on a compact 26-muscle model~\cite{yuan2026smat,yuan2026smatvalidation}. Here we ask two further questions. Can co-adaptive training remain stable, and transfer to real users, when the human is a far richer full-body model with a population of body sizes and gaits? And how does physiological energetics, with and without explicit mechanical-power shaping, influence the assistance learned through human--exoskeleton co-adaptation, and can the resulting policy transfer to human users?

This study addresses both questions with MARL-Exo, a multi-agent framework built on a full-body model with 304 Hill-type muscle actuators. Its human policy outputs residual joint targets adapted to a reference gait trajectory, which a learned muscle-activation layer converts into muscle activations. Its exoskeleton policy sees only intentionally delayed hip kinematics and its own recent commands, so that it can run on inexpensive wearable sensing. Training uses a two-stage schedule (human-only gait, then joint co-adaptation), and the exoskeleton reward includes the metabolic power of the hip muscles, evaluated with a muscle-energetics model, either alone (Metabolic) or together with explicit mechanical-power shaping (Metabolic\,+\,Power). The Metabolic policy was deployed on the same hardware and eight-participant cohort of the SMAT evaluation~\cite{yuan2026smatvalidation}. The contributions are as follows:
\begin{enumerate}
\item the use of a physiologically based muscle-energetics model as a term of the exoskeleton control policy's reward, within a two-stage multi-agent framework that couples a full-body human model and a hip exoskeleton, through regularized constraints and conditions the policy on randomized body and gait parameters;
\item a simulation comparison showing that, with the metabolic term retained, removing explicit positive-power shaping lowered modeled metabolic power across the analyzed muscle set by 13.8\,\% relative to no-device walking, against 8.8\,\% when power shaping was included; and
\item a human validation in eight healthy adults,  in which the deployed policy lowered net metabolic rate by $18.3\,\%$ relative to the passive device in all participants (adjusted $p<0.001$), more than offset the device's $13.1\,\%$ wearing penalty in six of eight participants (mean $7.4\,\%$ and up to $24.2\,\%$ below no-device walking), enlarged hip range of motion by $36.8\,\%$, and delivered predominantly positive estimated assistance power ($98.5\,\%$).
\end{enumerate}
The rest of this paper is organized as follows. Section~\ref{sec:marl} describes the simulation model and training framework, Section~\ref{sec:human} the hardware and human experiment, Section~\ref{sec:results} the simulation and experimental results, and Sections~\ref{sec:discussion} and~\ref{sec:conclusion} discuss and conclude.

\section{Cooperative Multi-Agent Training}
\label{sec:marl}
\subsection{Human--Exoskeleton Coupling}

The human and the hip exoskeleton are modeled as two separate multibody
systems in a shared physical simulation world simulated with the DART engine~\cite{lee2018dart}; each has its own segment, joint, and inertia definitions. The human musculoskeletal model has 50 joint degrees of freedom (DOFs) (excluding the six global or root DOFs) and is driven by $304$ Hill-type muscle actuators (line-of-action elements, several of which can represent one anatomical muscle)~\cite{park2022generative,park2023bidirectional}. The hip exoskeleton weighs 5.94\,kg and has 4 DOFs, including two passive abduction/adduction DOFs and two actuated flexion/extension DOFs.
The hip exoskeleton is rigidly attached to the human model
through three permanent bilateral constraints resolved by the constraint
solver: a weld fusing the pelvis to the exo backpack (all six DOFs
locked) and a ball joint pinning each femur to the corresponding exo thigh
segment (the three translational degrees of freedom locked). The constraints are \emph{regularized} rather than ideally rigid: DART
enforces them through finite constraint-force-mixing and error-reduction
parameters, so the interface carries solver-level compliance while
remaining much stiffer and cheaper to solve than explicit elastic or bushing connections~\cite{zhou2021design, luo2023robust}.
Because the ball joints leave hip rotation free, the wearer's hip degrees of
freedom are not over-constrained; we expect the regularization to affect mainly transient
load transmission rather than the net sagittal-plane assistive
torque. Exoskeleton motor torques (peak at
$\pm\SI{18}{\newton\meter}$) act on the device's hip joints and are transmitted
to the human pelvis and thighs through these constraints, altering the muscle
effort required to produce the gait. 

\subsection{Reinforcement Learning Framework}
\label{subsec:framework}

\begin{figure*}[t]
    \centering
    \includegraphics[width=\textwidth]{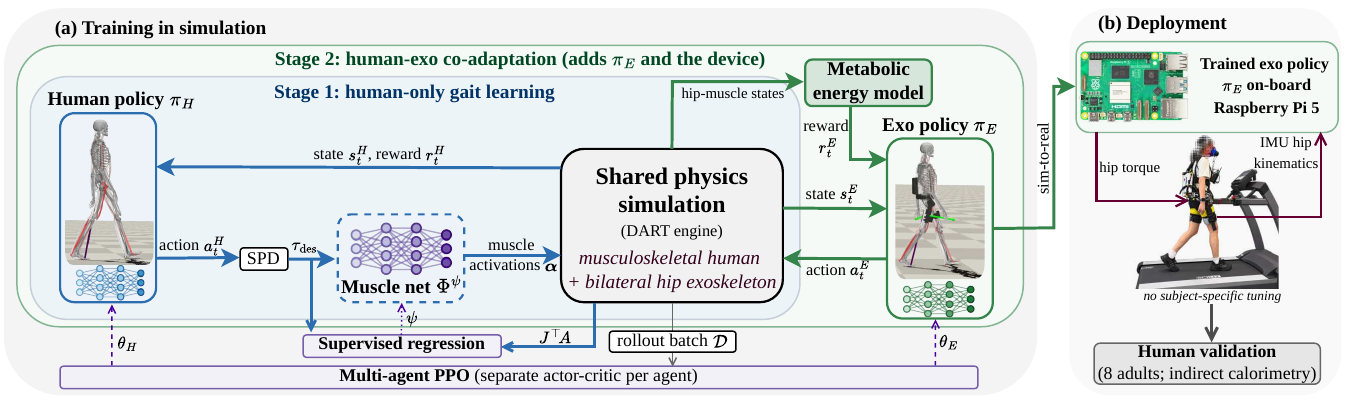}
\caption{Overview of MARL-Exo. (a) Training in simulation: a human policy $\pi_H$ and an exoskeleton policy $\pi_E$ act on a shared physics simulation; human actions pass through stable proportional-derivative (SPD) control and an auxiliary muscle network $\Phi^{\psi}$, and the exoskeleton reward is computed by a muscle metabolic energy model. Stage~1 trains $\pi_H$ alone, and Stage~2 adds the exoskeleton and co-adapts both policies. Multi-agent PPO updates both policies from a shared rollout batch $\mathcal{D}$, while $\Phi^{\psi}$ is trained by supervised regression to reproduce the desired joint torques $\tau_{\mathrm{des}}$ through the muscle moment-arm mapping $J^\top A$ (line 22 of Algorithm 1). (b) The trained $\pi_E$ is deployed on a Raspberry~Pi~5 without subject-specific tuning and validated in eight adults.}
    \label{fig:framework}
\end{figure*}

Hip assistance is posed as a multi-agent reinforcement learning (MARL)
problem~\cite{tan2025myoassist,lowe2017multi} with two
cooperating agents that share one simulated world (Fig.~\ref{fig:framework}): a
human policy $\pi_H$ that drives the musculoskeletal model and an exoskeleton
policy $\pi_E$ that commands hip torque. Each agent has its own observation,
reward, value function (critic), and PPO update~\cite{schulman2017proximal}, and the
two learn from one shared batch of rollouts. This framework shares its multi-agent formulation with
SMAT~\cite{yuan2026smat} but differs from it in the human model (304 rather than 26 muscle actuators, with randomized
anatomy and gait), the action pathway, the curriculum (two stages rather than
four), the exoskeleton reward, and separate agent-specific rewards and critics (Table~\ref{tab:smat_diff}).

\begin{table}[t]
\centering
\caption{Main differences between MARL-Exo (this work) and SMAT~\cite{yuan2026smat,yuan2026smatvalidation}.}
\label{tab:smat_diff}
\setlength{\tabcolsep}{3pt}
\footnotesize
\begin{tabular}{p{0.24\columnwidth}p{0.34\columnwidth}p{0.34\columnwidth}}
\toprule
 & SMAT & MARL-Exo \\
\midrule
Human model & 26 muscles (13 per side), imitation reward & 50-DOF body, 304 Hill-type muscle actuators, gait reward \\
Human action & Muscle excitations & Residual joint targets on a reference trajectory $\to$ PD torque $\to$ learned muscle network \\
Randomization & None & Anatomy, strength, gait, speed in observation \\
Curriculum & Four stages & Two stages \\
Critic & Shared by both actors & One per agent \\
Exo reward & Power sign and smoothness & Smoothness, magnitude, hip-muscle metabolic power, and weak human-reward coupling (deployed policy) \\
Online filtering & None & 3\,Hz angle and velocity \\
Experimental data & \multicolumn{2}{p{0.68\columnwidth}}{Same cohort and session; rest, No-Exo, and Passive recordings shared; active trials order randomized} \\
\bottomrule
\end{tabular}
\end{table}

\textbf{Observations.}
The human observation $s_t^H$ is a full-body proprioceptive vector:
the center-of-mass (COM) height; each body node's COM position relative to the
whole-body COM ($\in\mathbb{R}^3$) and orientation (first two rows of its world
rotation matrix, $\in\mathbb{R}^6$); each body node's linear velocity relative
to the COM velocity, the COM linear velocity, and each angular velocity (scaled
by $0.1$); the gait phase $\phi_t$ (with the local phase appended when the
time-warp is active); the longitudinal COM-to-next-foothold distance; the scaled
joint state (current and target angles, $\times 0.008$); and the sampled or randomized anatomical parameters (gait, skeleton-scale, and muscle-force factors) within their sampled ranges. These parameters include whole-body and per-segment skeleton scales ($\pm10\%$ global,
$\pm5\%$ per segment), the maximum isometric force of each muscle ($\pm25\%$), and gait stride and cadence ($\pm25\%$).
During training, the target walking speed was sampled from 0.65\,m/s to 2.0\,m/s, and the height and weight of the human model were randomized between [53.6\,kg, 1.55\,m] and [97.1\,kg, 1.94\,m].
Since these sampled parameters are appended to the policy's
observation, the controller is explicitly conditioned on the body and gait it drives rather than trained to be invariant. This parameter-conditioned domain randomization enables a single training to produce appropriate gaits across the sampled population. The resulting human observation has 649 dimensions, and the human policy outputs 51 actions: 50 residual joint-target offsets and one phase time-warp term, low-pass filtered with a 6\,Hz Butterworth filter at the 50\,Hz control rate. The residuals are added to a reference trajectory consisting of a single symmetrized, height-calibrated level-walking cycle
which acts as a motion prior; no motion-imitation reward is used.

The exo observation $s_t^E$ is compact and delayed: for
each of the two hip motors it stacks the joint angle and angular
velocity over the last $n_{\text{hist}}{=}3$ frames, delayed to
two control frames (i.e.\ 40\,ms) and low-pass filtered with a second-order Butterworth filter at 3\,Hz, concatenated with the last $n_{a}{=}3$ exo
actions (18 dimensions total). It outputs normalized bilateral hip torque commands
$\hat{\mathbf{u}}\in[-1,1]^{2}$, scaled by torque limits $\tau_{\max}$ to produce
physical assistance, through an MLP with hidden sizes $[128,64]$. In all exoskeleton training, an auxiliary symmetry loss (weight 10) added to the PPO objective penalized differences between the policy's output and its mirrored output for left--right mirrored observations.

\begin{algorithm}[t]
\caption{Cooperative MARL training of a human policy and a hip-exoskeleton
policy with an auxiliary muscle actuator net $\Phi^{\psi}$.}
\label{alg:marl}
\begin{algorithmic}[1]
\Require PPO hyperparameter configurations; iterations $I$; workers $N$;
muscle moment-arm matrix $J^\top\!A$;
exo motor torque limit $\tau_{\max}$; reference joint trajectory
$q^{\text{ref}}(\phi)$ (gait phase $\phi$) for the residual policy; action scale $c_a=0.04$
\State Init human policy $\pi_H^{\theta_H}$ and exo policy $\pi_E^{\theta_E}$
\State Init muscle net $\Phi^{\psi}$; broadcast $\psi$ to all workers
\For{$i = 1$ \textbf{to} $I$}
  \ParFor{worker $n = 1$ \textbf{to} $N$}
    \While{episode not terminated}
      \State $s_t^H \gets \Call{getHumanState}{}$
      \State $s_t^E \gets \Call{getExoState}{}$
      \State $a_t^H \sim \pi_H(\cdot\,|\,s_t^H)$;
             $a_t^E \sim \pi_E(\cdot\,|\,s_t^E)$
      \State target angles
             $q_t^{\text{target}} \gets q_t^{\text{ref}}(\phi) + c_a\, a_t^H$
      \State desired human joint torque
             $\tau_{\text{des}} \gets \Call{SPD}{q_t^{\text{target}}}$
      \State muscle activations
             $\boldsymbol{\alpha} \gets \Phi^{\psi}(J^\top\!A,\ \tau_{\text{des}})$
             \Comment{applied each substep}
      \State exo torque $\tau_{E} \gets a_t^E \times \tau_{\max}$
      \State \Call{env.step}{}
             \Comment{advance one control step $(\phi)$ ($n_{\text{sub}}$ physics substeps, and advance phase$(\phi)$)}
      \State $r_t^H \gets \Call{getHumanReward}{}$
      \State $r_t^E \gets \Call{getExoReward}{}$
      \State store $(s_t^{H},a_t^{H},r_t^{H})$, $(s_t^{E},a_t^{E},r_t^{E})$
             in $\mathcal{D}$; cache $(J^\top\!A,\ \tau_{\text{des}})$
      \State terminate on end-of-episode / NaN
    \EndWhile
  \EndParFor
  \State compute per-agent GAE advantages from $\mathcal{D}$
  \State $\theta_H \gets \theta_H + \eta\,\nabla_{\theta_H} J_H^{\text{clip}}$;\quad
         $\theta_E \gets \theta_E + \eta\,\nabla_{\theta_E} J_E^{\text{clip}}$
  \Statex \quad\textit{// Auxiliary supervised muscle learning}
  \State $\boldsymbol{\alpha} \gets \Phi^{\psi}(J^\top\!A,\ \tau_{\text{des}})$;\quad
         $\psi \gets \arg\min_{\psi}\,
         \lVert (J^\top\!A)\,\boldsymbol{\alpha} - \tau_{\text{des}} \rVert^2
         + \lambda \lVert \boldsymbol{\alpha} \rVert^2$ (Adam); broadcast $\psi$
  \State $\bar r^{H},\ \bar r^{E} \gets$ mean episode reward of the human
         and exo agent over the rollouts of iteration $i$
\EndFor
\State \Return $\pi_H^{\theta_H},\ \pi_E^{\theta_E},\ \Phi^{\psi}$
\end{algorithmic}
\end{algorithm}

\subsection{Two-Stage Training}
\label{subsec:curriculum}

Training both agents jointly from scratch can often become unstable, as the
exoskeleton optimizes against a human policy that has not yet acquired
stable gait.
We therefore separate training into two
sequential stages: locomotion is first learned without the device, and
the mechanically coupled exoskeleton is introduced only in the second
stage, where both policies are optimized jointly. Algorithm~\ref{alg:marl} describes the training loop used in both stages; the stages differ in the presence of the exoskeleton and in the active reward terms.

\textbf{Human Baseline Gait Learning.}
Stage~1 trains the human policy alone, without the device, to walk stably at varied speeds, and the resulting policy is the starting point for co-adaptation.
With a gait-reward gain $w_g{=}2$, the human reward multiplies locomotion,
velocity, and foot-placement terms and adds an activation-based effort term plus auxiliary
regularizers:
\begin{equation}
r_t^H = w_g\, r_{\text{loco}}\, r_{\text{avg}}\, r_{\text{step}}
        + w_{\text{eff}}\, r_{\text{eff}} + \textstyle\sum_k r_{\text{aux}}^{k},
\end{equation}
\begin{equation}       
\begin{aligned}
r_{\text{loco}} &= e^{-4\lVert\Delta \dot{x}_{\text{head}}\rVert^2}\;
                   e^{-4\,\Delta\theta_{\text{head}}^{2}}, &
r_{\text{avg}}  &= e^{-6\lVert \bar v_{\text{com}} - v^{*}\hat z \rVert^2}, \\
r_{\text{step}} &= e^{-2\lVert \Delta_{\text{foot}}\rVert^2 / n}, &
r_{\text{eff}}  &= e^{-8\,\overline{\lVert \boldsymbol{\alpha}\rVert^2}/N_m},
\end{aligned}
\end{equation}
where $\Delta\dot{x}_{\text{head}}$ and $\Delta\theta_{\text{head}}$ are the head
linear-acceleration and orientation deviations, $\bar v_{\text{com}}$ is the current COM velocity, 
$v^{*}=\ell_{\text{stride}}/
T_{\text{stride}}$ is the target COM speed, $\hat z$ is the forward direction vector, 
$\Delta_{\text{foot}}$ is the
dead-zone-clipped swing-foot error, $\overline{\lVert\boldsymbol{\alpha}\rVert^2}$
is the mean squared muscle activation over the control interval
($N_m$ muscles), and $w_{\text{eff}}{=}0.2$. The term $r_{\text{eff}}$ is an activation proxy for human effort and is distinct from the physiological energetics term of the exoskeleton reward below. The auxiliary terms
$r_{\text{aux}}^{k}$ penalize joint-limit violations and insufficient foot clearance and reward smooth joint torques (formulas and constants in Supplementary Section~I).

\textbf{Human--Exoskeleton Co-Adaptation.}
This stage begins the joint training of the human and exoskeleton policies. The
human control policy is loaded from the human baseline training. 

The exo reward sums an action-smoothness penalty, an action-magnitude penalty,
an optional joint-power term, a muscle metabolic-power term, and a small coupling to the human
reward:
\begin{equation}
\begin{aligned}
r_t^E ={} & -\big(0.2\lVert\Delta a^E\rVert^2 + \lVert\Delta^2 a^E\rVert^2\big)
            - 0.01\lVert a^E\rVert^2 \\
          & {}+ w_{\text{p}}\, r_{\text{power}} + r_{\text{energy}} + 0.02\, r_t^H,
\end{aligned}
\end{equation}
\begin{equation}
\begin{gathered}
r_{\text{power}} = \sum_j\!\big[\max(0,P_j) - 5\max(0,-P_j)\big], \\
r_{\text{energy}} = \max\!\Big(1 - \frac{\dot E_{\text{Uchida}}}{\dot E_0},\ -5\Big),
\end{gathered}
\end{equation}
where $\Delta a^E,\ \Delta^2 a^E$ are the first/second temporal differences of the
exo action, $P_j = \tau_j^{E}\,\dot q_j$ is the mechanical power at hip motor $j$
(so positive/assistive power is rewarded and negative/resistive power penalized
$5\times$), and $\dot E_{\text{Uchida}}$ (W) is the Uchida-model metabolic power of the
hip muscles detailed below, normalized by a constant $\dot E_0 = 150$\,W; the floor at $-5$ (i.e., $\dot E_{\text{Uchida}} \ge 6\dot E_0$) guards against extreme values caused by simulation instabilities. Two exoskeleton controllers were trained whose rewards differ only in $w_{\text{p}}$:
\emph{Metabolic\,+\,Power} ($w_{\text{p}}=0.005$) and \emph{Metabolic} ($w_{\text{p}}=0$). The names refer to the terms being compared, not the entire objective; both retain the smoothness, magnitude, metabolic, and human-coupling terms. Both controllers were obtained within the co-adaptation stage. The \emph{Metabolic\,+\,Power} controller was co-adapted from the Stage-1 human policy until its exoskeleton reward plateaued.
The Metabolic controller was then initialized from the converged best checkpoint and co-adaptation continued with the power term removed,
so the deployed policy is a reward refinement within Stage~2 rather than an independently trained controller. The Metabolic controller is the one deployed on hardware. 

\textbf{Physiologically Based Metabolic Energy Model.}
$\dot E_{\text{Uchida}}$ follows the muscle-energetics model of Umberger \emph{et al.}~\cite{umberger2003model} as implemented by Uchida \emph{et al.}~\cite{uchida2016stretching}, summed over the muscle actuators that span the hip flexion/extension degree of freedom of either leg, 
i.e., all actuators with a nonzero moment arm about that axis. In the present model these are 46 actuators per side (92 in total), representing adductor brevis, longus, and magnus, gluteus maximus, medius, and minimus, iliacus, psoas major, pectineus, tensor fasciae latae, sartorius, gracilis, rectus femoris, the hamstrings (biceps femoris long head, semimembranosus, semitendinosus), and the short external rotators (piriformis, gemelli, obturators, quadratus femoris). Each muscle's metabolic rate is the sum
of a heat rate and a mechanical work rate, scaled by the mass $m_i$ of muscle $i$:
\begin{equation}
\begin{gathered}
\dot E_{\text{Uchida}} = \sum_{i} \big(\dot H_i + \dot W_i\big)\, m_i, \\
\dot H_i = \max\!\big(\dot H_{\text{AM},i} + \dot H_{\text{S},i},\ 1\big),
\end{gathered}
\label{e_dot_Uchida}
\end{equation}
where $\dot H_i$ (in W/kg, floored at $1$) combines the activation--maintenance
heat $\dot H_{\text{AM}}$ and the shortening/lengthening heat $\dot H_{\text{S}}$, and $\dot W$ is
the mechanical work rate. With muscle activation $a\in[0,1]$,
active force--length factor $F_{\text{iso}}$, normalized fiber velocity $\tilde v = v_{ce}/\ell_m^0$,
where $v_{ce}$ is the muscle lengthening rate,
$\ell_m^0$ is the optimal muscle fiber length, and aerobic scale $\varphi=1.5$, the components are (muscle index omitted)
\begin{equation}
\begin{gathered}
\dot H_{\text{AM}} = \varphi\, a^{0.6}\, \zeta(F_{\text{iso}}),
\quad
\dot H_{\text{S}} = \varphi\, a^{p}\, \sigma(\tilde v), \\
\dot W = -\frac{F^{\text{act}} v_{ce}}{m},
\end{gathered}
\label{asw_dot_Uchida}
\end{equation}
where $p{=}2$ for shortening ($\tilde v\!\le\!0$) and $p{=}1$ for lengthening,
$F^{\text{act}} = a\, f_0\, F_{\text{iso}}$ is the active fiber force ($f_0$: maximum isometric force),
and $\zeta,\sigma$ are the fiber-length and fiber-velocity heat scalings, which also depend on the recruited fiber-type composition. 

The PPO hyperparameters, shared by both agents and both stages, are listed in Table~\ref{tab:ppo_params}. The physics ran at 500\,Hz and both policies at 50\,Hz (10 physics substeps per control step), with human joint SPD (Stable Proportional Derivative~\cite{park2022generative}) gains $k_p=200$ and $k_v=40$. Training ran on a workstation with two 16-core Intel Xeon Gold 5218 CPUs (2.3\,GHz) and an NVIDIA Quadro RTX 5000 GPU, using 16 rollout workers; the co-adaptation runs took about 24\,h for 62.6\,M control steps for the \emph{Metabolic\,+\,Power} training and 14\,h for 38.9\,M control steps for the \emph{Metabolic} training until convergence.

\begin{table}[t]
\centering
\caption{PPO training hyperparameters. The same
configuration is shared by the human policy $\pi_H$ and the exoskeleton
policy $\pi_E$; the auxiliary muscle network is trained by supervised
regression.}
\label{tab:ppo_params}
\setlength{\tabcolsep}{4pt}
\begin{tabular}{lc}
\toprule
Parameter & Value \\
\midrule
\multicolumn{2}{l}{\textit{Policy optimization (PPO, per agent)}} \\
Learning rate            & $1{\times}10^{-4}$ \\
PPO clip range $\epsilon$ & 0.2 \\
Value-function clip      & 100 \\
Value loss coefficient   & 1.0 \\
Discount $\gamma$        & 0.99 \\
GAE $\lambda$            & 0.99 \\
SGD iters per update     & 4 \\
Train batch size         & 8192 \\
Minibatch size           & 512 \\
KL coefficient (initial) & 0.01 \\
Target KL                & 0.01 \\
Entropy coefficient      & 0 \\
Gradient clipping        & none \\
Batch mode               & truncate episodes \\
Horizon                  & 10000 \\
Parallel workers ($1$ env each) & 16 \\
\addlinespace
\multicolumn{2}{l}{\textit{Auxiliary muscle network (supervised)}} \\
Learning rate            & $1{\times}10^{-4}$ \\
Epochs per update        & 10 \\
Minibatch size           & 512 \\
\bottomrule
\end{tabular}
\end{table}

\subsection{Simulation Evaluation}
Simulations used a full-body musculoskeletal model with a total body mass of 72.9\,kg. Walking was simulated at approximately 1.14\,m/s with a mean gait-cycle duration of ${\sim}$1.10\,s. Three conditions were analyzed: unassisted walking without the device (NoExo, a human-only model); an active hip exoskeleton whose controller was trained with both the positive power term $r_{\text{power}}$ and the metabolic term $r_{\text{energy}}$ (Metabolic\,+\,Power, labeled PowerExo); and an active hip exoskeleton whose controller was trained with $r_{\text{power}}$ removed while retaining $r_{\text{energy}}$ (Metabolic, labeled EnergyExo). The EnergyExo controller is the policy deployed on hardware and is referred to as MARL-Exo in the human experiment.

The analysis covers 31 muscle actuators per leg, representing 15 anatomical muscles of the hip, knee, and ankle. They were grouped by joint and direction into six functional groups (hip extension and flexion, knee flexion and extension, ankle plantar flexion and dorsiflexion) and by joint into three joint groups (hip, knee, and ankle, each summing both directions). Group membership is listed in Supplementary Section~II (Table~S2). Biarticular muscles belong to groups at more than one joint.

For each condition we computed muscle-contributed joint torques at the lower limb joints; cycle-averaged muscle activation; and metabolic power per muscle (Watts), aggregated by functional group and summed over all analyzed muscles (W and W/kg), reported per leg as the left--right average. Group activation is the unweighted mean over the member actuators, whereas group metabolic power is their sum. Each condition was evaluated in one deterministic episode at the default body and gait parameters with about 18
complete gait cycles. All quantities are phase-normalized to the gait cycle (0--100\%), reported bilaterally (left--right combined), and shown as the cycle mean with a $\pm$1\,SD band across cycles. 
The analyzed muscle set is sagittal-plane only, so absolute metabolic values underestimate whole-body cost. Because the same muscle set is used throughout, the conditions can be compared consistently on this partial-muscle total.

\section{Human Experiment}
\label{sec:human}

\subsection{Hip Exoskeleton Hardware and Deployment}
\label{subsec:hardware}

The trained MARL-Exo control policy was deployed to a custom bilateral hip exoskeleton, whose mechanical design, size adjustments, and full specifications are given in~\cite{yuan2026smatvalidation}; only the key elements needed are repeated here. The device weighs 5.94\,kg and provides one actuated sagittal flexion/extension degree of freedom per side (MyActuator X8-25, 25\,N$\cdot$m peak) together with a passive abduction/adduction hinge for fitting and lateral compliance.
Bilateral hip angle was taken as the pelvis--femur difference measured by three wireless IMUs (LPMS-B2, LP-Research, Tokyo, Japan) mounted on the pelvis and both thighs and zeroed in a standing posture before each trial, and hip angular velocity as the corresponding gyroscope difference about the flexion axis. The trained MARL-Exo policy ran on an onboard Raspberry~Pi~5 and reproduced the simulated observation and action interfaces exactly. At each 50\,Hz control step the actor consumed the 18-dimensional observation defined in Section~\ref{subsec:framework} and emitted a normalized torque in $[-1,1]$ per hip, which was scaled by the preset per-hip torque limit and transmitted to the actuators over the CAN bus. A second-order Butterworth low-pass filter with a 3\,Hz cutoff, designed for the 50\,Hz loop rate, was applied online to the measured hip angle and angular velocity, matching the observation filter used in training. Trials were run under a software torque limit and an emergency stop, with an operator supervising at all times.

\subsection{Human Validation Framework}
\label{subsec:protocol}

Eight healthy adults (5 male, 3 female; age $27.0\pm7.8$\,yr, range 18--44; mass $70.4\pm15.0$\,kg; height $1.73\pm0.06$\,m) participated under IRB approval (Protocol No.\ 2305033091R003) with written informed consent. These participants and this session also supplied the data of the SMAT evaluation~\cite{yuan2026smatvalidation}. The standing-rest, no-exoskeleton, and passive recordings are therefore common to both studies and are reused here as the reference conditions (Table~\ref{tab:metabolic} and Fig.~\ref{fig:metabolic}); the MARL-Exo trials and every analysis involving them are new.

\begin{figure*}[t]
\centering
\includegraphics[width=\textwidth]{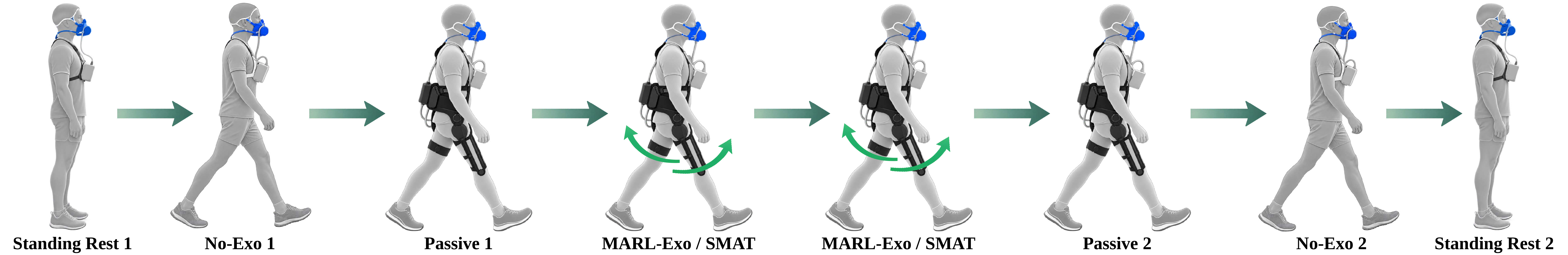}
\caption{Experimental protocol. Each participant completed standing rest, no-exoskeleton walking (No-Exo), and passive walking (Passive), then the two active controllers (MARL-Exo and SMAT) in randomized order, followed by the same reference conditions in reverse order. Each walking trial lasted approximately 6\,min on a treadmill at 1.25\,m/s, with metabolic rate measured by indirect calorimetry throughout.}
\label{fig:protocol}
\end{figure*}

Before data collection, each participant completed a treadmill familiarization session, walking for 5\,min with the device worn in passive mode (zero commanded torque) and for 10\,min under each of the two active controllers (MARL-Exo and SMAT), after which the fit was left unchanged for the remainder of the experiment. Data collection followed a symmetric protocol with randomized active-controller order, designed to separate the metabolic penalty of wearing the device from the benefit of its active assistance: standing rest $\to$ no-exoskeleton walking $\to$ passive walking $\to$ two active trials (MARL-Exo and SMAT, order randomized; MARL-Exo first in 3 of 8 participants) $\to$ passive walking $\to$ no-exoskeleton walking $\to$ standing rest (Fig.~\ref{fig:protocol}). The passive condition retained the full device mass with commanded torque held at zero and the actuators left back-drivable, isolating the mechanical burden of the hardware from the assistance it delivers.

Both active trials used identical hardware and differed only in the deployed policy. MARL-Exo is the controller referred to as Ctrl-B in~\cite{yuan2026smatvalidation}, where it was run in the same session but not analyzed. MARL-Exo was commanded at a 12\,N$\cdot$m torque limit. The SMAT policy saturates its normalized output near $0.8$ rather than unity~\cite{yuan2026smatvalidation} and was therefore commanded at 15\,N$\cdot$m, so that its realized peak ($0.8\times15\approx12$\,N$\cdot$m) matched that of MARL-Exo; both controllers thus operated at a common ${\sim}12$\,N$\cdot$m level of assistance. Metabolic and kinematic data were recorded identically in both active trials.

Placing the no-exoskeleton, passive, and standing-rest conditions both before and after the active block helps assess fatigue and drift over the session, and the two repeats of each were averaged before analysis; averaging reduces, but does not eliminate, the influence of nonlinear drift or carryover between active trials. Every walking trial lasted approximately 6\,min on a treadmill at 1.25\,m/s, and expired gases were sampled breath-by-breath with a portable indirect calorimetry system (COSMED K5, COSMED, Italy). Between conditions, participants rested seated for at least 5\,min.

Metabolic rate was computed from breath-by-breath oxygen uptake $\dot V_{O_2}$ and carbon-dioxide output $\dot V_{CO_2}$ (mL/min) with the Brockway relation~\cite{brockway1987derivation}.
The respiratory exchange ratio (RER, $\dot V_{CO_2}/\dot V_{O_2}$) was checked in every walking trial, with values below 1 confirming predominantly aerobic metabolism. Each condition was represented by the mean of the last 2\,min of its trial, the two repeats of the No-Exo, Passive, and rest conditions were averaged, and subtracting the standing-rest value gave the net rate $\dot E_{\text{net}}$. The contrasts of interest are the device penalty (Passive vs.\ No-Exo) and the assistance benefit (MARL-Exo vs.\ Passive and No-Exo).

\subsection{Biomechanical and Statistical Analysis}
\label{subsec:analysis}

Hip angle, angular velocity, and commanded torque were logged at 50\,Hz, and analysis used a steady-state window of the last 2\,min of each trial. 
The logged velocity was refiltered offline with a zero-phase fourth-order Butterworth low-pass at 6\,Hz, which adds no further phase delay; it cannot, however, remove any delay already present in a causally filtered logged signal (Section~\ref{subsec:hardware}). 
Cycles were segmented at peak hip flexion, since no ground-reaction-force sensors were used.

Hip range of motion (ROM) was the difference between peak flexion and peak extension in each cycle. Cadence was computed from the mean cycle duration, and stride length was derived from it at the fixed belt speed, so the two are reciprocal by construction. Kinematic metrics were computed per cycle, then averaged within each subject and across the two hips.

Mechanical output was summarized with the metrics of~\cite{lim2023parametric}, also used in~\cite{yuan2026smatvalidation}: RMS torque ($\tau_\text{RMS}$), mean peak torque ($\tau_\text{MAX}^\text{flex}$, $\tau_\text{MAX}^\text{ext}$), and the mean positive and negative power over the $n$ samples of a cycle,
\begin{equation}
\text{MPP} = \frac{1}{n}\!\sum_{\tau_i\dot{q}_i>0}\!\tau_i\dot{q}_i,
\quad
\text{MNP} = \frac{1}{n}\!\sum_{\tau_i\dot{q}_i<0}\!\tau_i\dot{q}_i,
\label{eq:mpp_mnp}
\end{equation}
with $P=\tau\dot q$ the estimated assistance power, computed from the commanded torque and the IMU-derived hip angular velocity converted to rad/s. 
The positive-power ratio $R_\text{pos}=\text{MPP}/(\text{MPP}+|\text{MNP}|)$ is the positive estimated work divided by the total absolute estimated work over the same interval; it describes the timing of assistance relative to hip motion and is not a measure of metabolic efficiency. These were averaged like the kinematic metrics.

Percentage changes in net metabolic rate were computed per subject and then averaged, and comparisons used two-tailed paired $t$-tests ($n=8$) after Shapiro--Wilk screening of the paired differences, Bonferroni-corrected within each outcome family at $\alpha=0.05$. The metabolic family has three planned comparisons (Passive vs.\ No-Exo, MARL-Exo vs.\ Passive, MARL-Exo vs.\ No-Exo), and the kinematic family has six (MARL-Exo vs.\ No-Exo for hip ROM, peak flexion, peak extension, cadence, and stride length, and Passive vs.\ No-Exo for hip ROM). Confidence intervals are unadjusted paired 95\,\% intervals. 
The participant was the inferential unit throughout; gait cycles were averaged within participants before testing.

\section{Results}
\label{sec:results}

\subsection{Simulation Evaluation Results}
\label{subsec:res_sim}

\subsubsection{Hip joint kinematics and kinetics}
\begin{figure}[!t]
\centering
\includegraphics[width=\columnwidth]{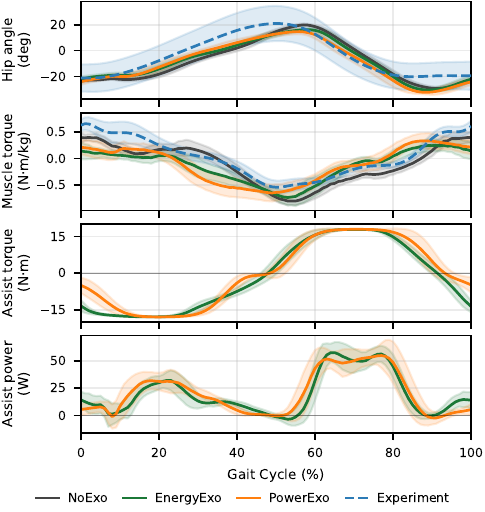}
\caption{Hip joint angles, muscle-contributed joint moment (mass-normalized), assistance torque, and assistance power over the gait cycle for the simulated conditions (PowerExo: Metabolic\,+\,Power; EnergyExo: Metabolic) and an independent biological reference (Experiment, teal dashed, mean $\pm$ 1\,SD)~\cite{camargo2021comprehensive}. Muscle-contributed moments are not identical to inverse-dynamics net moments.}
\label{fig:hip_data}
\end{figure}

In the simulation evaluation, both active controllers (PowerExo and EnergyExo) reduced peak hip extension relative to NoExo, with PowerExo producing the larger reduction, and slightly deepened peak hip flexion (most under PowerExo), consistent with the device delivering a flexion assist during swing initiation (Fig.~\ref{fig:hip_data}). 
Overlaying the biological reference (Fig.~\ref{fig:hip_data}) shows qualitative agreement in hip kinematics: the simulated hip trajectory and range follow the biological curve. 
At the hip, both controllers reduced the peak flexion torque, with EnergyExo producing the larger reduction (Fig.~\ref{fig:hip_data}); hip-torque range was reduced by both. 
Plotted in mass-normalized units, the simulated joint moments agree with the biological reference in overall shape and timing at the hip joint: both show the hip extensor-to-flexor transition through stance.
Fig.~\ref{fig:hip_data} also shows the gait-cycle-averaged assistance torque and power.
Compared to the PowerExo assistance, the EnergyExo assistance provided slightly higher $\tau_{\text{RMS}}$ (13.62 vs.\ 13.12\,N$\cdot$m) but slightly lower MPP (19.98 vs.\ 20.64\,W) with $\tau_{\max}=18$\,N$\cdot$m, both computed from the phase-averaged assistance curve in the figure.

\subsubsection{Muscle activation and metabolic energy by functional group}
Total cycle-averaged metabolic power summed over the analyzed muscles fell under both controllers, with EnergyExo achieving the larger reduction ($-$13.8\%, 1.830 $\to$ 1.578\,W/kg per leg) and PowerExo an intermediate reduction ($-$8.8\%, $\to$ 1.670\,W/kg per leg) (Table~\ref{tab:energy}). Summed by joint, the savings were concentrated proximally: metabolic power of the hip muscles fell by 29.8\% under EnergyExo and 16.0\% under PowerExo, that of the knee muscles by 12.8\% and 11.9\%, and that of the ankle muscles by only 3.9\% and 2.5\%. The two controllers therefore differed mainly at the hip, where EnergyExo used 4.2\,W less than PowerExo, whereas their knee and ankle totals differed by less than 1.3\,W. Because biarticular muscles are counted at every joint they span, the joint rows overlap, and each controller's hip and knee savings together exceed its total reduction.
Within each joint, the savings were unevenly distributed between directions. At the hip, both directions contributed, with larger reductions in Hip Extension (EnergyExo $-$35.4\%, PowerExo $-$19.2\%) than in Hip Flexion ($-$24.0\%, $-$12.6\%). At the knee, the reduction came almost entirely from Knee Extension ($-$36.2\%, $-$26.5\%), while Knee Flexion changed little ($-$4.3\%, $-$6.6\%). At the ankle, the plantar-flexor group was preserved near baseline ($-$1.7\%, $+$0.6\%), and the small ankle reduction came mainly from the dorsiflexor ($-$12.7\%, $-$14.9\%). Both controllers thus secured broad proximal savings while leaving ankle plantar-flexor energy essentially unchanged.

\begin{figure*}[!t]
\centering
\includegraphics[width=\textwidth]{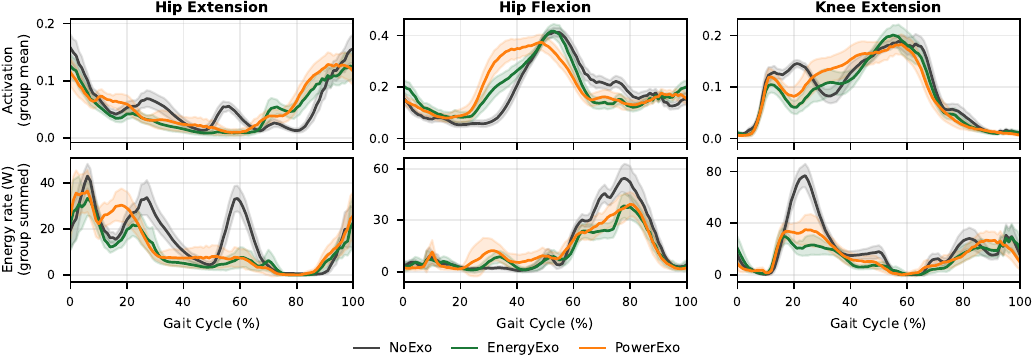}
\caption{Muscle activation (top row) and metabolic energy rate (bottom row) over the gait cycle for the three functional groups where the controllers produce their largest reductions: Hip Extension, Hip Flexion, and Knee Extension (group mean activation; group summed energy; mean $\pm$ 1\,SD). Hip-flexor activation rises during stance phase but falls during swing phase while hip-flexor energy mostly falls under both controllers.}
\label{fig:merged}
\end{figure*}

The activation curves (Fig.~\ref{fig:merged}, top row) show that both controllers reduce hip-extensor and knee-extensor activation while hip-flexor activation rises and falls during the gait cycle. The energy curves (Fig.~\ref{fig:merged}, bottom row) show that all three groups lose metabolic energy under both controllers, with EnergyExo generally below PowerExo. The hip flexors thus show an activation--energy divergence: group mean activation rises while summed group energy falls ($-$24.0\%, $-$12.6\%; Table~\ref{tab:energy}). Because activation is an unweighted group mean, whereas metabolic power is a sum weighted by muscle mass (Eq.~\ref{e_dot_Uchida}; see also~\cite{leem2026exo}), this divergence may partly reflect aggregation and redistribution among member muscles. 

\begin{table*}[!t]
\centering
\caption{Cycle-averaged metabolic power by functional group (sum across member muscles) and total across all involved muscle actuators, all values per leg (left--right averaged; a partial-muscle sum, not whole-body metabolism). PowerExo: Metabolic\,+\,Power reward; EnergyExo: Metabolic reward (deployed). $\Delta$ in Watts; percentages signed vs.\ NoExo. Groups overlap through biarticular muscles, so the six group totals exceed the total; the TOTAL row is the true non-overlapping sum. The Hip, Knee, and Ankle rows sum both directions at each joint (no muscle belongs to both directions of the same joint); biarticular muscles still appear in more than one joint row. Group membership is given in Supplementary Table~S2. W/kg is the per-leg power divided by body mass (72.9\,kg).}
\label{tab:energy}
\newcommand{\B}[1]{\textbf{\boldmath #1}}
\setlength{\tabcolsep}{10pt}
\small
\begin{tabular}{lcccccc}
\toprule
Muscle group & NoExo & PowerExo & EnergyExo & NoExo  & PowerExo$-$NoExo & EnergyExo$-$NoExo \\
&  W &  W &  W &  W/kg & $\Delta$ (\%) & $\Delta$ (\%) \\
\midrule
Hip Extension & 15.47 & 12.50 & 9.99 & 0.212 & $-$2.97 ($-$19.2\%) & $-$5.48 ($-$35.4\%) \\
Hip Flexion & 14.84 & 12.97 & 11.27 & 0.204 & $-$1.87 ($-$12.6\%) & $-$3.57 ($-$24.0\%) \\
Knee Flexion & 56.07 & 52.39 & 53.64 & 0.769 & $-$3.68 ($-$6.6\%) & $-$2.43 ($-$4.3\%) \\
Knee Extension & 20.32 & 14.94 & 12.97 & 0.279 & $-$5.38 ($-$26.5\%) & $-$7.35 ($-$36.2\%) \\
Ankle PlantarFlex & 69.60 & 70.02 & 68.40 & 0.955 & $+$0.42 ($+$0.6\%) & $-$1.20 ($-$1.7\%) \\
Ankle DorsiFlex & 17.31 & 14.73 & 15.12 & 0.237 & $-$2.58 ($-$14.9\%) & $-$2.19 ($-$12.7\%) \\
\midrule
Hip (Ext.\,+\,Flex.) & 30.30 & 25.47 & 21.26 & 0.416 & $-$4.84 ($-$16.0\%) & $-$9.04 ($-$29.8\%) \\
Knee (Flex.\,+\,Ext.) & 76.40 & 67.34 & 66.61 & 1.048 & $-$9.06 ($-$11.9\%) & $-$9.79 ($-$12.8\%) \\
Ankle (PF\,+\,DF) & 86.91 & 84.75 & 83.52 & 1.192 & $-$2.16 ($-$2.5\%) & $-$3.39 ($-$3.9\%) \\
\midrule
\B{TOTAL} & \B{133.43} & \B{121.75} & \B{115.01} & \B{1.830} & \B{$-$11.68 ($-$8.8\%)} & \B{$-$18.42 ($-$13.8\%)} \\
\bottomrule
\end{tabular}
\end{table*}

\subsection{Metabolic Cost of Walking from Experiment}
\label{subsec:res_metabolic}

\begin{figure}[!t]
\centering
\includegraphics[width=\columnwidth]{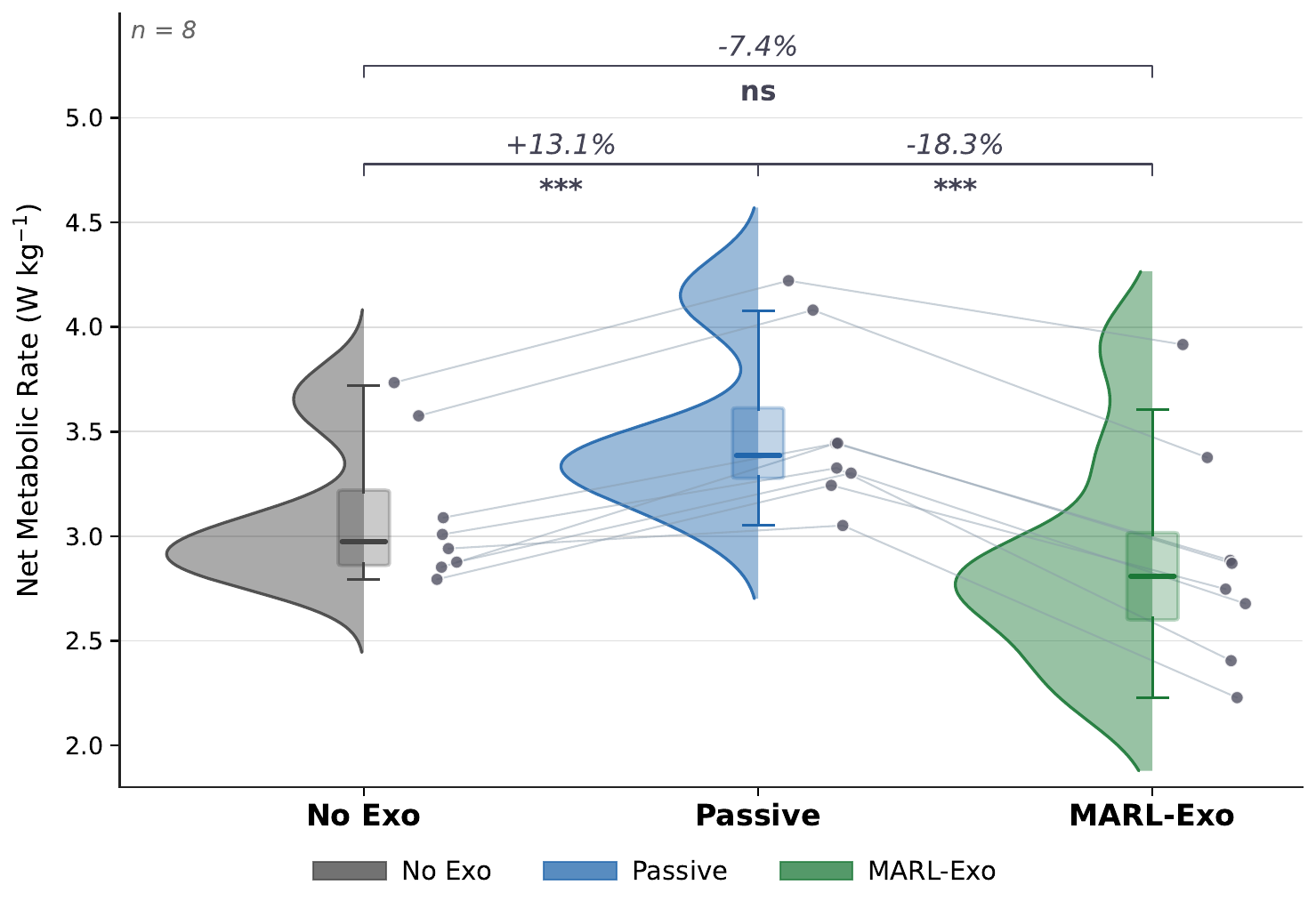}
\caption{Net metabolic rate (W\,kg$^{-1}$) across conditions ($n=8$). Half-violin distributions, box plots, and individual subjects (connected lines) for No-Exo, Passive, and MARL-Exo. The No-Exo and Passive data are shared with~\cite{yuan2026smatvalidation}. Brackets show the group-mean percentage change between conditions ($+13.1\,\%$, $-18.3\,\%$, $-7.4\,\%$). Paired $t$-test with Bonferroni correction: $^{*}p<0.05$, $^{**}p<0.01$, $^{***}p<0.001$; n.s.\ $p\ge0.05$. All eight subjects reduced net cost under assistance relative to the passive condition.}
\label{fig:metabolic}
\end{figure}

Net metabolic rate across the three conditions is summarized in Fig.~\ref{fig:metabolic} and Table~\ref{tab:metabolic}. The RER, averaged over the same final 2\,min of each trial, remained below unity in all 40 analyzed walking trials (five per participant: two No-Exo, two Passive, and one MARL-Exo; overall $0.838\pm0.032$, range $0.785$--$0.916$; No-Exo: $0.831\pm0.035$, Passive: $0.839\pm0.030$, MARL-Exo: $0.847\pm0.029$), so the Brockway relation was applied throughout within its aerobic range. The two repeats of the no-exoskeleton and passive conditions differed by $6.9\pm5.2\,\%$ and $3.5\pm2.7\,\%$ of gross metabolic rate respectively, with no consistent direction across subjects, so we found no evidence of a systematic trend over the session; averaging the repeats reduces, but does not remove, this variability.

Carrying the unpowered device raised net metabolic rate by $13.1\pm4.9\,\%$ above unassisted walking ($t_7=7.80$, corrected $p<0.001$), a penalty consistent with the added mass and the constraint the structure imposes, although the present data do not separate these causes. This penalized condition, rather than unassisted walking, isolates the contribution of the assistance itself.

Against that baseline, active MARL-Exo assistance lowered net metabolic rate by $18.3\pm6.6\,\%$, from $3.514\pm0.414$ to $2.888\pm0.538$\,W/kg ($t_7=-9.22$, corrected $p<0.001$). The effect was present in every participant, with individual reductions spanning $7.2\,\%$ to $27.7\,\%$, so the assistance more than recovered the cost of wearing the hardware.

Relative to walking without the device, six of the eight participants walked at a lower net cost with the assisted device than without any device, with individual reductions of up to $24.2\,\%$ and above $20\,\%$ in two participants; in these six, the assistance recovered between $1.1$ and $7.5$ times the metabolic penalty of wearing the hardware. Across the cohort the mean reduction was $7.4\pm10.0\,\%$ (paired difference $-0.221$\,W/kg, unadjusted 95\,\% CI $[-0.474,\,+0.033]$, $t_7=-2.05$, corrected $p=0.24$), with the interval reflecting the between-subject spread in an eight-participant cohort.

\begin{table*}[t]
\caption{Per-subject net metabolic rate and reductions under MARL-Exo and SMAT ($n=8$).}
\label{tab:metabolic}
\centering\footnotesize
\renewcommand{\arraystretch}{0.9}
\begin{tabular*}{\textwidth}{@{\extracolsep{\fill}}lcccccccc@{}}
\toprule
 & \multicolumn{4}{c}{Net metabolic rate (W/kg)} & \multicolumn{2}{c}{MARL-Exo reduction (\%)} & \multicolumn{2}{c}{SMAT reduction (\%)} \\
\cmidrule(lr){2-5}\cmidrule(lr){6-7}\cmidrule(l){8-9}
Subject & No-Exo & Passive & MARL-Exo & SMAT & vs Passive & vs No-Exo & vs Passive & vs No-Exo \\
\midrule
S1 & 3.008 & 3.326 & 2.405 & 2.738 & 27.7 & 20.1 & 17.7 & 9.0 \\
S2 & 3.575 & 4.081 & 3.376 & 3.221 & 17.3 & 5.6 & 21.1 & 9.9 \\
S3 & 2.941 & 3.051 & 2.228 & 2.368 & 27.0 & 24.2 & 22.4 & 19.5 \\
S4 & 2.794 & 3.243 & 2.747 & 2.711 & 15.3 & 1.7 & 16.4 & 3.0 \\
S5 & 3.734 & 4.222 & 3.916 & 3.571 & 7.2 & $-$4.9 & 15.4 & 4.4 \\
S6 & 2.876 & 3.301 & 2.678 & 2.458 & 18.9 & 6.9 & 25.5 & 14.5 \\
S7 & 2.852 & 3.444 & 2.884 & 2.802 & 16.3 & $-$1.1 & 18.6 & 1.8 \\
S8 & 3.088 & 3.445 & 2.870 & 2.744 & 16.7 & 7.1 & 20.3 & 11.2 \\
\midrule
Mean\,$\pm$\,SD & 3.11\,$\pm$\,0.35 & 3.51\,$\pm$\,0.41 & 2.89\,$\pm$\,0.54 & 2.83\,$\pm$\,0.39 & 18.3\,$\pm$\,6.6 & 7.4\,$\pm$\,10.0 & 19.7\,$\pm$\,3.3 & 9.1\,$\pm$\,6.0 \\
\bottomrule
\end{tabular*}\\[2pt]
\parbox{\textwidth}{\scriptsize Reduction $=(\dot E_\text{ref}-\dot E_\text{active})/\dot E_\text{ref}$, positive when the active condition costs less, computed per subject and then averaged. The SMAT data are from~\cite{yuan2026smatvalidation}.}
\end{table*}

\subsection{Hip Biomechanics and Assistance Power Delivery}
\label{subsec:res_biomech}

\begin{figure*}[t]
\centering
\includegraphics[width=\textwidth]{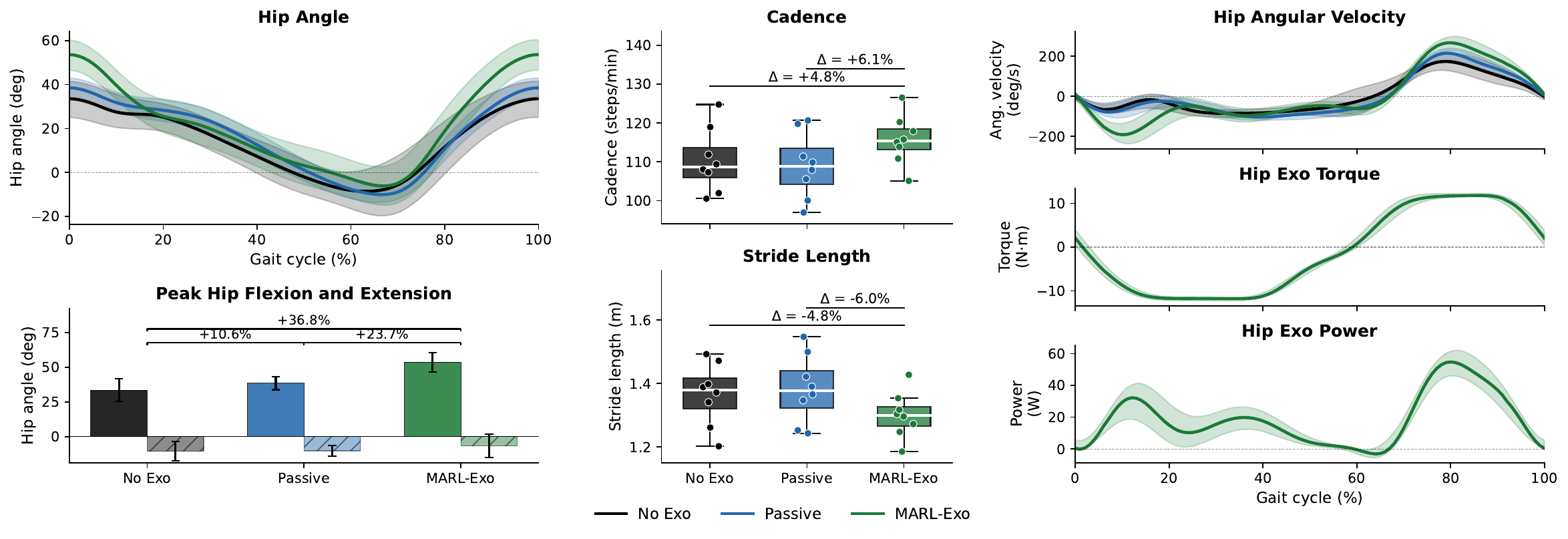}
\caption{Treadmill walking at 1.25\,m/s under No-Exo (black), Passive (blue), and MARL-Exo (green) ($n=8$; per-subject means of both hips). \emph{Left:} gait-cycle hip angle (flexion positive) with peak flexion and extension; labels give the change in ROM. \emph{Middle:} cadence and stride length. \emph{Right:} hip angular velocity together with the exoskeleton torque and mechanical power delivered by MARL-Exo. Bands and error bars show mean $\pm$1\,SD across subjects. The No-Exo and Passive data are shared with~\cite{yuan2026smatvalidation}.}
\label{fig:biomech}
\end{figure*}

Fig.~\ref{fig:biomech} compares hip kinematics and delivered assistance power across the No-Exo, Passive, and MARL-Exo conditions. Hip kinematics were derived from the three body-mounted IMUs, whose standing-posture zero offsets each subject's absolute joint angles. Range of motion is taken as the primary kinematic outcome, 
while the individual flexion and extension peaks are reported as descriptive quantities. The paired differences satisfied the Shapiro--Wilk normality check for every variable except peak flexion ($W=0.809$, $p=0.036$), for which a Wilcoxon signed-rank test is reported alongside the parametric result.

Active assistance markedly enlarged hip range of motion, from $44.0\pm4.3^\circ$ without the device to $60.2\pm6.6^\circ$ under MARL-Exo, an increase of $36.8\,\%$ ($t_7=5.97$, corrected $p=0.003$), with the passive condition falling between the two ($48.7\pm4.0^\circ$, $+10.6\,\%$, $t_7=3.93$, corrected $p=0.034$). The enlargement was carried almost entirely by the flexion phase: peak flexion rose by $20.1^\circ$, from $33.6\pm8.2^\circ$ to $53.6\pm6.9^\circ$, a $59.8\,\%$ increase ($t_7=5.38$, corrected $p=0.006$), whereas peak extension was statistically unchanged ($-10.5\pm6.9^\circ$ to $-6.6\pm8.4^\circ$, $t_7=1.19$, corrected $p>0.99$). Both range of motion and peak flexion increased in all eight participants without exception. For peak flexion, whose paired differences were not normally distributed, the nonparametric test agreed in direction and remained significant after correction ($W=0$, exact $p=0.008$, corrected $p=0.047$); with $n=8$ the exact test cannot resolve below $p=0.008$, so this value reflects the sample size rather than the magnitude of the effect, which the unanimous direction and the $20.1^\circ$ shift characterize more directly.

The spatiotemporal structure of the gait shifted far less than its amplitude. Cadence rose from $110.3\pm8.2$ to $115.7\pm6.3$\,steps/min under assistance, a $4.8\,\%$ increase that did not survive correction ($t_7=2.92$, corrected $p=0.13$); at the fixed belt speed this is equivalent to a shortening of stride length from $1.366\pm0.098$ to $1.300\pm0.072$\,m, the two being reciprocal by construction. Assistance thus amplified hip excursion, while the changes in cadence and stride length were smaller and not significant after correction.

The exoskeleton delivered torque in both directions of the stride, with a flexion-assist peak in late swing and an extension-assist phase through mid-stance. Table~\ref{tab:mech} reports the per-subject mechanical output. Mean per-cycle peak torque was $11.79\pm0.39$\,N$\cdot$m in flexion and $-11.81\pm0.42$\,N$\cdot$m in extension, magnitudes that did not differ significantly ($t_7=1.07$, $p=0.32$), indicating that the policy saturated symmetrically against the $12$\,N$\cdot$m commanded limit in both directions; RMS torque was $9.45\pm0.45$\,N$\cdot$m. Mean positive estimated power was $19.19\pm2.74$\,W while mean negative estimated power remained near zero ($-0.30\pm0.14$\,W), giving a positive-power ratio $R_\text{pos}$ of $0.985\pm0.006$.
Because the deployed policy was trained without the explicit power term, this predominantly positive power emerged from the metabolic and regularization terms rather than being directly rewarded. 
The spread across subjects was small in all four metrics, indicating that the single deployed policy produced consistent mechanical assistance without subject-specific tuning.

\begin{table}[t]
\caption{Per-subject exoskeleton mechanical output, MARL-Exo condition ($n=8$, both hips averaged).}
\label{tab:mech}
\centering \setlength{\tabcolsep}{4pt}
\resizebox{\columnwidth}{!}{%
\begin{tabular}{lcccccc}
\toprule
Subject & $\tau_\text{RMS}$ (N$\cdot$m) & $\tau_\text{MAX}^\text{flex}$ (N$\cdot$m) & $\tau_\text{MAX}^\text{ext}$ (N$\cdot$m) & MPP (W) & MNP (W) & $R_\text{pos}$ \\
\midrule
S1 & 9.62 & 11.94 & $-$11.97 & 24.04 & $-$0.54 & 0.978 \\
S2 & 9.76 & 11.97 & $-$11.98 & 21.02 & $-$0.24 & 0.989 \\
S3 & 9.36 & 11.92 & $-$11.94 & 16.79 & $-$0.20 & 0.988 \\
S4 & 10.10 & 11.99 & $-$11.99 & 21.53 & $-$0.33 & 0.985 \\
S5 & 9.42 & 11.97 & $-$11.97 & 18.40 & $-$0.39 & 0.979 \\
S6 & 9.68 & 11.80 & $-$11.96 & 18.26 & $-$0.05 & 0.997 \\
S7 & 8.75 & 10.84 & $-$10.78 & 16.16 & $-$0.29 & 0.982 \\
S8 & 8.89 & 11.86 & $-$11.88 & 17.29 & $-$0.33 & 0.981 \\
\midrule
Mean\,$\pm$\,SD & 9.45\,$\pm$\,0.45 & 11.79\,$\pm$\,0.39 & $-$11.81\,$\pm$\,0.42 & 19.19\,$\pm$\,2.74 & $-$0.30\,$\pm$\,0.14 & 0.985\,$\pm$\,0.006 \\
\bottomrule
\end{tabular}}
\vspace{0.5mm}
\begin{minipage}{\columnwidth}\footnotesize
$\tau_\text{RMS}$: RMS torque; $\tau_\text{MAX}^\text{flex}$, $\tau_\text{MAX}^\text{ext}$: mean per-cycle peak torque in the flexion- and extension-assist directions; MPP/MNP: mean positive/negative power (Eq.~\ref{eq:mpp_mnp}); $R_\text{pos} = \text{MPP}/(\text{MPP}+|\text{MNP}|)$: positive-power ratio. All metrics were computed per gait cycle and averaged across cycles within each subject, with the two hips averaged.
\end{minipage}
\end{table}

\section{Discussion}
\label{sec:discussion}

A central design choice of this work is to include a physiologically detailed metabolic model in the exoskeleton reward. In simulation, with the metabolic term retained in both controllers, removing explicit positive-power shaping lowered modeled metabolic power by a further $5$ percentage points ($-13.8\,\%$ vs.\ $-8.8\,\%$) and changed where the savings fell. Further training of the Metabolic\,+\,Power controller over a comparable number of steps did not improve it, which points to the reward change, not the additional training, as the source of the gain. Power shaping thus steers the learned strategy away from energetically cheaper solutions that the metabolic term alone finds. The divergence of hip-flexor activation and energy noted above suggests a mechanism: an energetics-based reward can favor strategies that an activation-based reward would penalize. Because energy models disagree with one another~\cite{miller2014comparison}, the simulated ranking of controllers and the savings by muscle group are specific to the Uchida model used here and may differ under alternatives such as that of Bhargava \emph{et al.}~\cite{bhargava2004phenomenological}. The reward otherwise follows trajectory-optimization practice, where energy is one term of a composite cost because energy alone can predict unrealistic movement and its rate expressions are non-smooth~\cite{ackermann2010optimality,koelewijn2018metabolic}.

On hardware, the central result is that assistance beat the unpowered device in every one of the eight participants ($-18.3\pm6.6\,\%$, $p<0.001$), with reductions from $7.2\,\%$ to $27.7\,\%$, even though the policy was neither retrained nor tuned for any of them. Because the passive condition carries the same hardware, this contrast isolates what the controller itself contributes, and we treat it as the primary outcome. On a common scale, active assistance removed $0.63$\,W/kg relative to the passive device, about $1.5$ times the $0.41$\,W/kg that wearing the unpowered device added. The net result relative to walking without any device therefore depends on the hardware as much as on the controller. Six of eight participants walked at a lower net cost than without the device, two by more than $20\,\%$, and the cohort mean was $-7.4\pm10.0\,\%$.

The savings relative to no-device walking reported for other simulation-trained hip controllers, $24.3\,\%$~\cite{luo2024experiment} and $15.2\,\%$~\cite{barati2026end}, combine the benefit of each controller with the penalty of its particular device. Isolated from the hardware by the passive condition, the benefit of MARL-Exo ($18.3\,\%$) exceeds the no-device saving reported in~\cite{barati2026end}, and it was achieved under less favorable hardware conditions than either study. Our device weighs $5.94$\,kg, about twice the $3.2$\,kg~\cite{luo2024experiment} and $2.9$\,kg~\cite{barati2026end} of those devices, and it was operated at a $12$\,N$\cdot$m limit, lower than the peak torque reported in~\cite{luo2024experiment}. The smaller no-device saving is therefore consistent with the $13.1\,\%$ penalty of our hardware rather than with an inefficient controller. 
Accordingly, a lighter device and a higher torque limit would be expected to shift the no-device contrast in favor of assistance.

The delivered assistance is consistent with this outcome. Torque was applied in both directions of the stride, with matching peaks in flexion and extension, and almost all of the estimated assistance power acted with the wearer's hip motion (Table~\ref{tab:mech}), the mechanical signature of well-timed assistance. Hip range of motion grew markedly, almost entirely through flexion, while cadence and stride length did not change significantly after correction. This pattern is consistent with an assistance profile that helps swing initiation, and with the simulated reduction of hip-flexor and hip-extensor energy. The larger excursion at the same belt speed is an altered gait response, so part of the metabolic benefit may come from the changed gait as well as from reduced muscle workload.

Because the two deployed controllers (SMAT and MARL-Exo) were tested in the same participants and session at a matched $\sim$12\,N$\cdot$m peak, they can be compared directly. Relative to the passive device, SMAT lowered net metabolic rate by $19.7\,\%$ and MARL-Exo by $18.3\,\%$, and per subject MARL-Exo was on average $0.061$\,W/kg higher in net cost (unadjusted 95\,\% CI $[-0.116,\,+0.239]$\,W/kg; $t_7=0.82$, uncorrected $p=0.44$). Both controllers thus reduced cost relative to the passive device by similar amounts, and this exploratory comparison detected no difference between them. The mechanical pattern was also nearly identical: flexion assistance in late swing and extension assistance through stance, the same positive-power ratio ($0.985$ vs.\ $0.984$), similar mean positive power ($19.2$ vs.\ $18.1$\,W), and similar gains in hip range of motion ($+36.8\,\%$ vs.\ $+40.8\,\%$). MARL-Exo had a slightly higher RMS torque ($9.45$ vs.\ $8.84$\,N$\cdot$m) with a slightly lower mean peak ($11.79$ vs.\ $12.39$\,N$\cdot$m), which suggests a flatter profile held near the limit for longer.

The similar metabolic benefit may partly reflect the shared hardware limit. Both controllers drove the same device, saturated against the same $\sim$12\,N$\cdot$m cap, and delivered similar positive mechanical power, and in a rigid hip exoskeleton the metabolic saving is expected to scale with the positive work returned to the wearer. Under a common torque ceiling, two policies that both learn to time assistance with hip motion may therefore converge on similar benefit, and the flatter, higher-RMS profile of MARL-Exo did not lower cost further, as a broad optimum or a ceiling would predict. Testing both controllers at higher torque limits, on lighter hardware, and at more than one speed would show where their differences in training become visible.

Several sim-to-real details deserve mention. The observation was built from IMU-derived hip angle and velocity, so it did not depend on motor encoders, and it was low-pass filtered online at 3\,Hz, as in training. 
The policy also ran at 50\,Hz on a Raspberry~Pi~5, and the observation latency modeled in simulation ($40$\,ms) is what allows such a modest processor to be used. Soft-tissue compliance and strap looseness still make IMU-derived angles an approximation of true hip motion, and the phase alignment of assistance might be improved by more accurate thigh and pelvis sensing~\cite{seel2014imu}. Finally, hardware was commanded at a $12$\,N$\cdot$m limit, lower than the $\pm18$\,N$\cdot$m simulated peak, to keep the device comfortable and to avoid wobble and excessive relative movement between the exoskeleton and the wearer, which the rigidly constrained simulation does not capture. The deployed assistance is therefore weaker than the one that produced the simulated savings, and higher torque with a better-fitting interface may yield larger benefit.

The scope of the study defines the next steps. The human study evaluated eight healthy adults walking on a treadmill at a single speed after 10\,min of familiarization with each controller, with reference conditions shared with our SMAT study. Larger cohorts would resolve the contrast with unassisted walking (corrected $p=0.24$ here), longer or multi-day exposure would show whether the benefit grows as wearers adapt to the assistance, and testing across speeds, slopes, and gait-impaired populations would exercise the body and gait randomization already present in training. Adding EMG and full lower-limb motion capture to the hip kinematics recorded here would allow the simulated redistribution of muscle effort, and the gait changes that accompany assistance, to be checked in humans.
Finally, a lighter device with a better-fitting interface would reduce the measured wearing penalty and allow the higher torque used in training to be applied without relative motion at the cuffs.

\section{Conclusion}
\label{sec:conclusion}
We presented MARL-Exo, a two-stage multi-agent reinforcement learning framework in which a full-body musculoskeletal human policy and a hip-exoskeleton policy are co-adapted in simulation, with a physiologically based muscle-energetics term in the exoskeleton reward. Co-adaptation remained stable with a 304-actuator human model and randomized anatomy, gait, and speed, and the choice of reward shaped the learned strategy: with the metabolic term retained, omitting explicit positive-power shaping lowered modeled muscle metabolic power by $13.8\,\%$ rather than $8.8\,\%$, with the additional savings concentrated at the hip and knee. 
The learned policy was deployed on a hip exoskeleton in eight healthy adults, without per-user tuning, and it reduced net metabolic rate by $18.3\pm6.6\,\%$ relative to the passive device, enlarged hip range of motion by $36.8\,\%$, and delivered almost entirely positive estimated assistance power. Six of eight participants walked below their no-device cost, by up to $24.2\,\%$, with a cohort-mean reduction of $7.4\pm10.0\,\%$. The benefit was close to that of our earlier SMAT controller at the same torque limit, despite a different human model, curriculum, and reward.
These results show that physiologically informed co-adaptive training can produce an exoskeleton controller that transfers from simulation to human users without per-user tuning, and that candidate reward designs can be compared in simulation before human testing. Larger and more diverse cohorts, lighter hardware, and evaluation across speeds, slopes, and gait-impaired populations are the next steps toward population-level assistance.

\section*{Acknowledgment}
This work was partially supported by the National Institute on Disability, Independent Living, and Rehabilitation Research (NIDILRR) funded Rehabilitation Engineering Research Center Grant 90REGE0025-01-00 and NSF Award \#2524089.

\bibliographystyle{IEEEtran}
\bibliography{reference}

\clearpage
\onecolumn

\setcounter{section}{0}
\setcounter{table}{0}
\setcounter{figure}{0}
\setcounter{equation}{0}
\renewcommand{\theHsection}{supp.\arabic{section}}
\renewcommand{\theHtable}{supp.\arabic{table}}
\renewcommand{\theHfigure}{supp.\arabic{figure}}
\renewcommand{\theHequation}{supp.\arabic{equation}}
\renewcommand{\thetable}{S\arabic{table}}
\renewcommand{\thefigure}{S\arabic{figure}}
\renewcommand{\theequation}{S\arabic{equation}}
\renewcommand{\arraystretch}{1.1}

\begin{center}
{\LARGE\bfseries Supplementary Material}\\[6pt]
{\large Multi-Agent Reinforcement Learning of Co-Adaptive Hip Exoskeleton Control\\
(MARL-Exo) for Metabolic Energy Reduction}\\[6pt]
{Yifei Yuan, Jakob Wolf, Ghaith Androwis, and Xianlian Zhou}
\end{center}
\vspace{1.5em}

\section{Auxiliary Terms of the Human Reward}
\label{sec:supp_aux}
The human reward contains several auxiliary terms
$\sum_k r_{\text{aux}}^{k}$ that regularize the learned gait. Three terms are
used, with identical settings in the human baseline stage and in both
co-adaptation runs (Metabolic and Metabolic\,+\,Power):
\begin{equation}
\sum_k r_{\text{aux}}^{k} = r_{\text{lim}} + r_{\tau} + r_{\text{clr}} .
\label{eq:supp_aux}
\end{equation}
Their constants are listed in Table~\ref{tab:supp_aux}.

\subsection{Knee Joint-Limit Penalty}
Knee hyperextension is penalized quadratically once the knee angle crosses a
soft limit placed a margin $\delta$ inside the joint's lower (extension) limit:
\begin{equation}
r_{\text{lim}} = -\,k_{\text{lim}} \sum_{j\in\{\text{L},\text{R}\}}
\max\!\big(0,\ (\theta_j^{\min}+\delta) - \theta_j\big)^{2},
\label{eq:supp_lim}
\end{equation}
where $\theta_j$ is the flexion angle of knee $j$ (flexion positive). With the
knee lower limit $\theta^{\min}=-0.1$\,rad and $\delta=0.1$\,rad, the penalty is
active for any hyperextension ($\theta_j<0$). The upper (flexion) limit is not
penalized.

\subsection{Joint-Torque Smoothness}
Smooth joint actuation is encouraged with a bounded term that decreases as the
desired joint torque changes more abruptly between control steps:
\begin{equation}
r_{\tau} = e^{-c\,\lVert\nabla\boldsymbol{\tau}_t\rVert^2}
 + 0.5\,e^{-c\,\lVert\nabla^2\boldsymbol{\tau}_t\rVert^2}
 + 0.2\,e^{-c\,\lVert\nabla^3\boldsymbol{\tau}_t\rVert^2},
\label{eq:supp_tau}
\end{equation}
where $\boldsymbol{\tau}_t$ is the vector of SPD desired joint torques
(N$\cdot$m) at control step $t$, $c=10^{-3}$, and $\nabla^n$ denotes the
$n$-th backward difference over consecutive control steps:
\begin{equation}
\begin{aligned}
\nabla\boldsymbol{\tau}_t &= \boldsymbol{\tau}_t-\boldsymbol{\tau}_{t-1}, \\
\nabla^2\boldsymbol{\tau}_t &= \boldsymbol{\tau}_t-2\boldsymbol{\tau}_{t-1}+\boldsymbol{\tau}_{t-2}, \\
\nabla^3\boldsymbol{\tau}_t &= \boldsymbol{\tau}_t-3\boldsymbol{\tau}_{t-1}+3\boldsymbol{\tau}_{t-2}-\boldsymbol{\tau}_{t-3}.
\end{aligned}
\label{eq:supp_diff}
\end{equation}
The term lies in $[0,\,1.7]$ and is applied once four control steps of torque
history are available.

\subsection{Foot-Clearance Penalty}
Foot scuffing is penalized when a reference point on the foot is below a
clearance height while moving horizontally:
\begin{equation}
r_{\text{clr}} = -\,k_{\text{clr}} \sum_{j\in\{\text{L},\text{R}\}}
\min\!\big(0,\ y_j - h_c\big)^{2}\,\big\lVert \mathbf{v}_j^{xz}\big\rVert,
\label{eq:supp_clr}
\end{equation}
where $y_j$ is the height of a fixed point on foot $j$ (\SI{0.15}{m} along the
foot segment from its origin), $\mathbf{v}_j^{xz}$ is that point's horizontal
velocity, and $h_c=0.03$\,m. The penalty is zero when the point is above $h_c$
or stationary on the ground, so it acts mainly during swing. For example, a point
\SI{1}{cm} below $h_c$ moving at \SI{1}{m/s} gives $r_{\text{clr}}=-0.01$.

\begin{table}[!ht]
\centering
\caption{Constants of the auxiliary human reward terms.}
\label{tab:supp_aux}
\setlength{\tabcolsep}{8pt}
\begin{tabular}{llll}
\toprule
Term & Symbol & Value & Description \\
\midrule
$r_{\text{lim}}$ & $k_{\text{lim}}$ & 100 & penalty gain \\
 & $\theta^{\min}$ & $-0.1$\,rad & knee lower (extension) limit \\
 & $\delta$ & 0.1\,rad & soft-limit margin \\
\addlinespace
$r_{\tau}$ & $c$ & $10^{-3}$\,(N$\cdot$m)$^{-2}$ & exponential scale \\
\addlinespace
$r_{\text{clr}}$ & $k_{\text{clr}}$ & 100\,s\,m$^{-3}$ & penalty gain \\
 & $h_c$ & 0.03\,m & clearance height \\
\bottomrule
\end{tabular}
\end{table}

\FloatBarrier
\section{Muscle Groups Used in the Simulation Analysis}
\label{sec:supp_groups}
The simulation analysis of the main text uses 31 muscle actuators per leg
(line-of-action elements of the 304-actuator model), representing 15 anatomical
muscles that act primarily in the sagittal plane. 
Each actuator is logged for the
left and right legs, and all reported values are left--right averages (per leg).
The actuators are assigned to six functional groups by joint and direction:
hip extension (HE), hip flexion (HF), knee flexion (KF), knee extension (KE),
ankle plantar flexion (PF), and ankle dorsiflexion (DF). Three joint groups
combine both directions at each joint: hip (HE\,+\,HF), knee (KF\,+\,KE), and
ankle (PF\,+\,DF). Table~\ref{tab:supp_groups} lists the membership.

No actuator belongs to both directions of the same joint, so each joint group is
the disjoint union of its two direction groups. Biarticular muscles (biceps femoris
long head, semimembranosus, semitendinosus, rectus femoris, and the gastrocnemius)
belong to groups at two joints, so the direction groups and the joint groups
overlap across joints and their sums exceed the total over all 31 actuators.
Group activation is the unweighted mean over member actuators, and group
metabolic power is their sum.

\begin{table}[!ht]
\centering
\caption{Membership of the analyzed muscle actuators (per leg) in the functional
and joint groups. $\bullet$ marks membership. The bottom rows give the number of
actuators in each group.}
\label{tab:supp_groups}
\setlength{\tabcolsep}{6pt}
\begin{tabular}{lc|cc|cc|cc}
\toprule
 & & \multicolumn{2}{c|}{Hip} & \multicolumn{2}{c|}{Knee} & \multicolumn{2}{c}{Ankle} \\
Anatomical muscle & Actuators & HE & HF & KF & KE & PF & DF \\
\midrule
Biceps femoris, long head   & 1 & $\bullet$ &           & $\bullet$ &           &           &           \\
Biceps femoris, short head  & 2 &           &           & $\bullet$ &           &           &           \\
Gastrocnemius, lateral head & 1 &           &           & $\bullet$ &           & $\bullet$ &           \\
Gastrocnemius, medial head  & 1 &           &           & $\bullet$ &           & $\bullet$ &           \\
Gluteus maximus             & 5 & $\bullet$ &           &           &           &           &           \\
Iliacus                     & 3 &           & $\bullet$ &           &           &           &           \\
Psoas major                 & 3 &           & $\bullet$ &           &           &           &           \\
Rectus femoris              & 2 &           & $\bullet$ &           & $\bullet$ &           &           \\
Semimembranosus             & 2 & $\bullet$ &           & $\bullet$ &           &           &           \\
Semitendinosus              & 1 & $\bullet$ &           & $\bullet$ &           &           &           \\
Soleus                      & 2 &           &           &           &           & $\bullet$ &           \\
Tibialis anterior           & 1 &           &           &           &           &           & $\bullet$ \\
Vastus intermedius          & 2 &           &           &           & $\bullet$ &           &           \\
Vastus lateralis            & 2 &           &           &           & $\bullet$ &           &           \\
Vastus medialis             & 3 &           &           &           & $\bullet$ &           &           \\
\midrule
Actuators per direction group & 31 & 9 & 8 & 8 & 9 & 4 & 1 \\
Actuators per joint group     &    & \multicolumn{2}{c|}{17} & \multicolumn{2}{c|}{17} & \multicolumn{2}{c}{5} \\
\bottomrule
\end{tabular}
\end{table}

\end{document}